\documentclass[pdflatex,sn-nature]{sn-jnl}

\usepackage{graphicx}%
\usepackage{multirow}%
\usepackage{amsmath,amssymb,amsfonts}%
\usepackage{amsthm}%
\usepackage[title]{appendix}%
\usepackage{xcolor}%
\usepackage{textcomp}%
\usepackage{manyfoot}%
\usepackage{booktabs}%
\usepackage{algorithm}%
\usepackage{algorithmicx}%
\usepackage{algpseudocode}%
\usepackage{listings}%
\usepackage{soul}
\usepackage{float}

\theoremstyle{thmstyleone}%
\theoremstyle{thmstyletwo}%

\theoremstyle{thmstylethree}%

\begin{document}

\title[Article Title]{Interpretable Fundus Image Classification via Ring-Based Retinal Vasculature Features}


\author[1]{\fnm{Xiaoyan} \sur{Li}}\email{xiaoy.li@mail.utoronto.ca}
\author[3]{\fnm{Shixin} \sur{Xu}}\email{shixin.xu@dukekunshan.edu.cn}
\author[1]{\fnm{Arvind} \sur{Gupta}}\email{arvind.gupta@utoronto.ca}
\author*[1,2,3,4]{\fnm{Huaxiong} \sur{Huang}}\email{hhuang@yorku.ca}

\affil*[1]{\orgdiv{Computer Science}, \orgname{University of Toronto}, \orgaddress{\street{27 King's College Cir.}, \city{Toronto}, \postcode{M5S 1A1}, \state{Ontario}, \country{Canada}}}

\affil[2]{\orgdiv{Mathematics and Statistics}, \orgname{York University}, \orgaddress{\street{4700 Keele St.}, \city{Toronto}, \postcode{M3J 1P3}, \state{Ontario}, \country{Canada}}}

\affil[3]{\orgdiv{Zu Chongzhi Center}, \orgname{Duke Kunshan University}, \orgaddress{\street{No.8 Duke Ave.}, \city{Kunshan}, \postcode{215300}, \state{Jiangsu}, \country{China}}}

\affil[4]{\orgdiv{Mathematics, Analytics, and Data Science Lab}, \orgname{Fields Institute for Research in Mathematical Sciences}, \orgaddress{\street{222 College Street}, \city{Toronto}, \postcode{M5T 3J1}, \state{Ontario}, \country{Canada}}}

\abstract{Retinal fundus photography is widely used for screening and monitoring ocular diseases, but many modern classification pipelines rely on deep latent representations and provide limited interpretability. This study develops an interpretable fundus image classification framework based on a ring-structured representation of the retinal vasculature centered on the optic disc. The method quantifies vessel geometry, color appearance, oxygenation-related vascular appearance, and vessel--background entropy within concentric retinal regions. These physiologically motivated descriptors are derived from vessel masks, image intensities, and optical-density measurements and aggregated across rings to capture spatial variation in vascular properties. Using only quantitative vascular descriptors, the proposed method achieved strong classification performance across three public fundus datasets. On HRF, it achieved 91.1\% accuracy using automatically generated vessel masks, matching RETFound, a vision transformer pretrained on large-scale retinal fundus image data, under the same evaluation setting. Additional analyses suggest that pretrained image models are sensitive to acquisition-related spatial cues, including fundus scale and retinal position within the field of view, as well as broader non-vessel image characteristics. This framework may support interpretable disease classification, quantitative retinal phenotyping, and retinal biomarker discovery without requiring large task-specific training datasets.}




\maketitle

\section{Introduction} \label{sec:introduction}

Diabetic retinopathy (DR) and glaucoma are among the leading causes of vision impairment and blindness worldwide~\cite{Grauslund2009ProliferativeDR,Klein2008WESDR25Year,Morizane2019VisualImpairmentJapan,Quigley2006GlaucomaWorldwide}. DR is a microvascular complication of diabetes and a major cause of vision loss in working-age adults, whereas glaucoma is a chronic optic neuropathy and a leading cause of irreversible blindness, particularly in aging populations. Identifying early retinal manifestations of these diseases is therefore critical for timely intervention and the prevention of vision loss.

Fundus photography represents one of the most widely used imaging modalities in ophthalmology and is routinely employed for screening, diagnosis, and longitudinal monitoring. Ophthalmologists visually assess color fundus images for disease-related signs, relying largely on qualitative judgment of vascular abnormalities (e.g., caliber changes, tortuosity), optic disc appearance, and the presence and distribution of lesions. Although fundus images contain rich structural, color, and textural information, qualitative grading may introduce subjectivity and variability. In this context, computer-aided diagnosis (CAD) systems can provide complementary quantitative assessments that are more reproducible across patients and imaging conditions.

Deep learning has driven substantial progress in automated fundus image analysis, achieving strong performance in DR and glaucoma classification~\cite{Akhtar2025DLDR,Wang2024ViTRetinal,Xu2024HybridDR,Bhoopalan2025TOViT,Vijayalakshmi2025MultiTaskDR,Chaurasia2025GlaucomaCV}. Vision transformer (ViT)--based architectures~\cite{Wang2024ViTRetinal,Bhoopalan2025TOViT} and hybrid CNN--ViT models~\cite{Xu2024HybridDR,Vijayalakshmi2025MultiTaskDR} have shown competitive performance relative to conventional convolutional approaches, particularly when supported by large-scale pretraining. However, despite their strong predictive performance, deep models typically rely on latent representations that are difficult to interpret in clinically meaningful terms. Their development may also require substantial training data and computational resources, and their predictions can be influenced by acquisition-related or other image characteristics that are not directly aligned with the retinal abnormalities used in clinical assessment.

From a clinical point of view, vascular alterations are central to the pathophysiology of both DR and glaucoma and are readily observable in color fundus images. DR is associated with microvascular remodeling, venous dilation, oxygenation-related appearance changes, and structural alterations in arterioles and venules. In glaucoma, peripapillary vascular attenuation and optic disc-centered changes have been documented~\cite{niemeijer2011avr,Cheung2010LancetDR,Weinreb2014JAMA}. These observations motivate the use of vascular descriptors as a clinically meaningful representation for fundus image classification. Moreover, because vascular morphology and appearance (including color- and oxygenation-related contrast) vary with retinal eccentricity (i.e., with distance from the optic disc), capturing these properties within concentric annular regions centered at the optic disc provides a structured, spatially organized characterization of retinal vasculature. Accordingly, there is a need for classification frameworks that retain strong discriminative power while providing a more transparent link between model predictions and clinically meaningful retinal features.

In this work, we introduce an interpretable ring-based vascular feature representation for fundus image classification. The proposed approach explicitly quantifies vascular geometry, color characteristics, oxygenation-related appearance, and vessel--background texture contrast within concentric annular regions centered on the optic disc. These measurements produce a compact set of physiologically motivated descriptors that capture spatial variations in vascular structure across retinal eccentricity. All features are computed from vessel masks and optical density measurements, after which ring-wise feature vectors are concatenated and used as input to a shallow classifier for three-class classification among healthy, DR, and glaucomatous fundus images.

Prior studies have investigated retinal vascular descriptors and oxygenation-related measurements, but these sources of information have often been examined separately. Our framework integrates oxygenation-related appearance with structural vascular descriptors within a unified ring-wise representation for interpretable classification across multiple retinal disease categories. Rather than focusing solely on absolute classification accuracy, we evaluate the discriminative information captured by explicit vascular descriptors relative to learned deep image representations. Accordingly, we compare the proposed method with pretrained image-based models, including RETFound~\cite{zhou2023foundation}, a ViT--based retinal foundation model pretrained on approximately 1.6 million retinal images and subsequently adapted to downstream classification tasks.

Furthermore, we examine how acquisition-related spatial characteristics, including field-of-view (FOV) size and retinal position within the image, influence the performance of pretrained image models. Although these characteristics may contribute to classification performance, they are not directly aligned with the retinal abnormalities commonly used in clinical assessment. In contrast, the proposed ring-based representation is constructed from physiologically meaningful vascular measurements, providing a more transparent link between model predictions and retinal characteristics.

The main contributions of this work are summarized as follows:
\begin{itemize}
    \item We propose an interpretable optic-disc-centered ring representation for fundus image classification that integrates vessel geometry, red--green color characteristics, oxygenation-related appearance, and vessel--background entropy across retinal eccentricity.

    \item We demonstrate across multiple public datasets that these compact, physiologically motivated descriptors provide strong classification performance and enable direct interpretation of feature and annular-region contributions, offering a transparent complement to pretrained deep image representations.

    \item We assess the sensitivity of pretrained image models to FOV size and retinal position within the image through controlled spatial standardization experiments. The observed performance changes highlight the potential influence of acquisition-related spatial characteristics when interpreting deep-model classification performance.
\end{itemize}

\section{Results}\label{sec:results}
\subsection{Benchmarking Against RETFound and Other Pretrained Image-Based Baselines}

Table~\ref{tab:retfound_vs_our} compares the proposed ring-based representation with RETFound~\cite{zhou2023foundation} and three ImageNet-pretrained image-based baselines: ResNet-50~\cite{he2016resnet}, ViT-B/16~\cite{dosovitskiy2021vit}, and ConvNeXt-B~\cite{liu2022convnext}. For the proposed method, results are reported using vessel masks predicted by RIP-AV~\cite{Dai2024} and, where available, expert vessel masks. Lightweight fine-tuning improved RETFound performance on HRF but did not yield consistent gains for the remaining datasets or pretrained backbones. We therefore report the fine-tuned RETFound result for HRF and frozen-feature results for the other evaluation settings.

On HRF, the proposed method achieved an accuracy of 1.000 using expert vessel masks. When predicted vessel masks were used, its accuracy decreased to 0.911, matching RETFound in accuracy and remaining higher than the other ImageNet-pretrained baselines. The difference between the ground-truth- and predicted-mask results indicates that vessel-segmentation quality can influence downstream classification, particularly for descriptors that depend on accurate delineation of fine vascular structures.

A progressive evaluation using ground-truth vessel masks further showed that HRF classification performance generally improved as additional peripheral annuli were included, particularly for the combined and geometry-based feature sets. This result suggests that peripheral vascular structure provides complementary information beyond the immediate peripapillary region. Detailed results are provided in Supplementary Section~S14, \textit{Progressive Ring-Wise Evaluation}.


On SUSTech-SYSU, all methods performed strongly for binary DR-versus-healthy classification. The proposed representation achieved an
accuracy of 0.896, while the pretrained image models achieved higher accuracies, with ConvNeXt-B and ViT-B/16 slightly outperforming RETFound. This suggests that, in this dataset, full-image models may have captured informative global cues beyond the explicit vascular measurements emphasized by the proposed method.

Performance was lower across methods on FIVES. Using expert vessel masks, the proposed method achieved an accuracy of 0.766, comparable to ConvNeXt-B at 0.769 and ViT-B/16 at 0.753, but approximately four percentage points below RETFound at 0.806. Its accuracy decreased to 0.724 when predicted vessel masks were used. The lower performance may partly reflect the greater image-quality variability in FIVES, including blur, distortion, and reduced contrast, which can obscure fine vascular structures and reduce the reliability of segmentation-dependent descriptors. Per-class sensitivity, specificity, F1-score, and one-vs-rest ROC-AUC for each dataset and method are reported in Supplementary Table~S7.

\begin{table}[!htbp]
\caption{Classification performance of the proposed method and pretrained image-based baselines.}
\label{tab:retfound_vs_our}%

\scriptsize
\renewcommand{\arraystretch}{1.05}

\begin{tabular*}{\textwidth}{@{\extracolsep\fill}llcccc@{}}
\toprule
\textbf{Dataset} & \textbf{Method}
& \textbf{Acc.}
& \textbf{Macro-F1}
& \textbf{ROC-AUC}
& \textbf{95\% CI} \\
\midrule

\multirow{6}{*}{HRF}
& Proposed (GT)
& \textbf{1.000}
& \textbf{1.000}
& \textbf{1.000}
& \textbf{[0.921, 1.000]} \\

& Proposed (Pred.)
& 0.911
& 0.910
& 0.964
& [0.793, 0.965] \\

& RETFound
& 0.911
& 0.911
& 0.966
& [0.793, 0.965] \\

& ViT-B/16
& 0.800
& 0.800
& 0.920
& [0.662, 0.891] \\

& ConvNeXt-B
& 0.867
& 0.868
& 0.958
& [0.738, 0.937] \\

& ResNet-50
& 0.778
& 0.780
& 0.901
& [0.637, 0.875] \\

\midrule

\multirow{6}{*}{FIVES}
& Proposed (GT)
& 0.766
& 0.740
& 0.884
& [0.721, 0.806] \\

& Proposed (Pred.)
& 0.724
& 0.700
& 0.843
& [0.677, 0.766] \\

& RETFound
& \textbf{0.806}
& \textbf{0.788}
& \textbf{0.931}
& \textbf{[0.763, 0.842]} \\

& ViT-B/16
& 0.753
& 0.741
& 0.879
& [0.708, 0.794] \\

& ConvNeXt-B
& 0.769
& 0.755
& 0.892
& [0.724, 0.809] \\

& ResNet-50
& 0.724
& 0.710
& 0.848
& [0.678, 0.767] \\

\midrule

\multirow{5}{*}{SYSU}
& Proposed (Pred.)
& 0.896
& 0.888
& 0.948
& [0.875, 0.913] \\

& RETFound
& 0.946
& 0.946
& 0.983
& [0.930, 0.958] \\

& ViT-B/16
& 0.953
& 0.949
& 0.977
& [0.938, 0.964] \\

& ConvNeXt-B
& \textbf{0.964}
& \textbf{0.961}
& \textbf{0.989}
& \textbf{[0.950, 0.974]} \\

& ResNet-50
& 0.944
& 0.940
& 0.977
& [0.928, 0.956] \\

\botrule
\end{tabular*}

\footnotetext{\textit{Note:}
HRF and FIVES were evaluated using leave-one-out cross-validation, whereas SYSU was evaluated using stratified five-fold cross-validation. Accuracy and its 95\% Wilson confidence interval were computed from aggregated out-of-fold predictions. Macro-F1 and ROC-AUC were also computed from the aggregated out-of-fold predictions. GT, ground-truth vessel masks; Pred., predicted vessel masks; SYSU, SUSTech-SYSU. For RETFound, the fine-tuned result is reported on HRF, whereas frozen-feature results are reported on FIVES and SYSU. On FIVES, all pretrained image-based baselines were evaluated using the FOV-standardized images described in Section~\ref{sec:acquisition_bases}.}
\end{table}

\subsection{Analysis of FIVES-Specific Acquisition-Related Spatial Characteristics}
\label{sec:acquisition_bases}

We investigated whether acquisition-related spatial characteristics in FIVES could influence the performance of pretrained image-based classifiers. Specifically, we examined FOV size, FOV center, and optic-disc location across disease groups. This analysis focused on FIVES because it showed the most pronounced group-wise spatial differences among the datasets studied, together with noticeable changes in classification performance after FOV standardization. The corresponding spatial differences were less pronounced in HRF and SUSTech-SYSU and did not produce comparably large changes in classification accuracy.

\begin{figure}[!htbp]
\centering

\begin{minipage}[t]{0.48\linewidth}
\centering
\includegraphics[width=\linewidth]{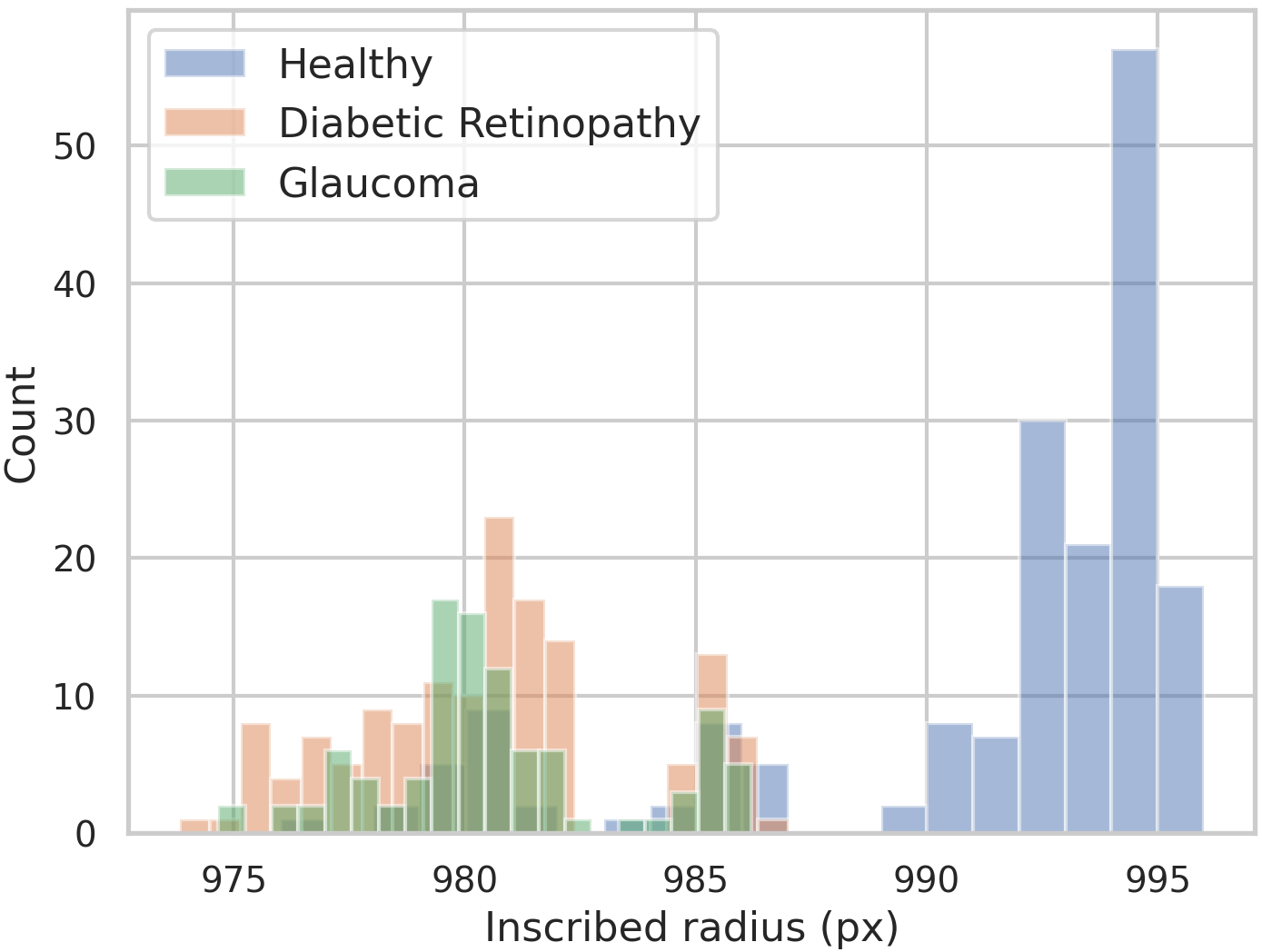}
\par\smallskip
\textbf{(a)}
\end{minipage}
\hfill
\begin{minipage}[t]{0.42\linewidth}
\centering
\includegraphics[width=\linewidth]{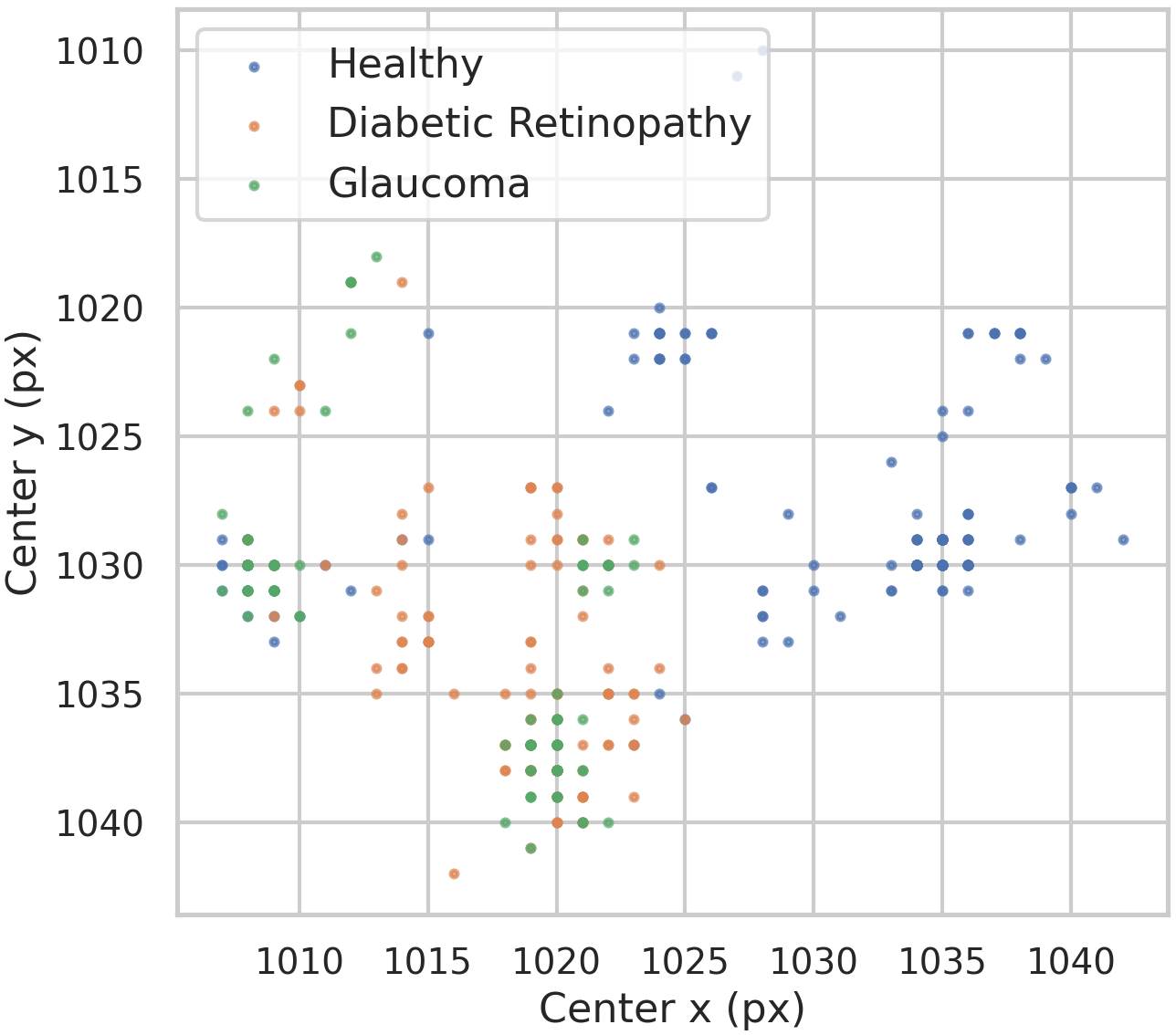}
\par\smallskip
\textbf{(b)}
\end{minipage}

\vspace{0.5em}

\begin{minipage}[t]{0.48\linewidth}
\centering
\includegraphics[width=\linewidth]{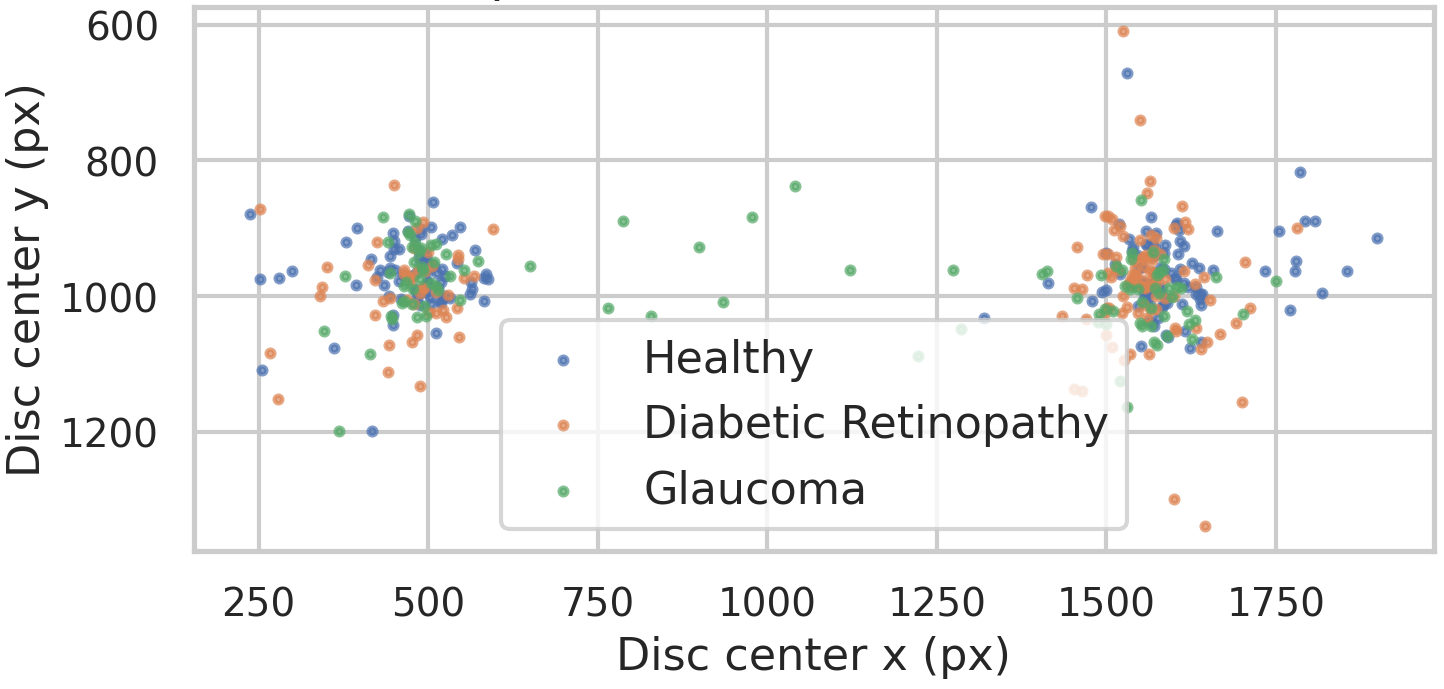}
\par\smallskip
\textbf{(c)}
\end{minipage}

\caption{Acquisition-related spatial characteristics in the FIVES dataset. (a) Distribution of the inscribed fundus FOV radius. (b) Spatial distribution of FOV centers. (c) Spatial distribution of optic-disc centers.}
\label{fig:fives_fov_analysis}
\end{figure}

As shown in Fig.~\ref{fig:fives_fov_analysis}, the disease groups exhibit distinct FOV-related spatial characteristics. Healthy images generally have larger inscribed fundus radii, concentrated around $990$--$996$ pixels, whereas glaucoma and DR images span a broader range of approximately $975$--$985$ pixels (Fig.~\ref{fig:fives_fov_analysis}(a)). Healthy images also show a more compact distribution of FOV centers, while glaucoma and DR images exhibit greater spatial variability (Fig.~\ref{fig:fives_fov_analysis}(b)). By contrast, the distributions of optic-disc centers largely overlap across groups, with no clear class-dependent separation (Fig.~\ref{fig:fives_fov_analysis}(c)). These observations suggest that the group-wise FOV differences are more likely related to image acquisition or preprocessing than to systematic anatomical variation.

To reduce the influence of these spatial characteristics, we standardized the FOV by retaining the common retinal region shared across all images. This procedure primarily affected the outer FOV boundary while preserving most of the valid retinal field. Nevertheless, classification performance decreased for all pretrained image-based models. RETFound showed the largest reduction in accuracy, from $0.927$ on the original images to $0.806$ after FOV standardization. Accuracy also decreased from $0.829$ to $0.769$ for ConvNeXt-B, from $0.819$ to $0.753$ for ViT-B/16, and from $0.798$ to $0.724$ for ResNet-50. These performance decreases suggest that the pretrained models were sensitive to FOV geometry, image boundaries, or related acquisition characteristics. Although this experiment does not isolate the contribution of each spatial cue, it demonstrates that acquisition-related spatial variation can materially affect image-based classification performance.




\subsection{Sensitivity of RETFound to Non-Vessel Background Appearance}
\label{sec:retfound_background_cues}

We further examined whether pretrained RETFound was sensitive to non-vessel image information beyond the vascular structures emphasized by the proposed representation. Using FOV-standardized FIVES images, we evaluated two complementary controls: broad perturbations that standardized or suppressed non-vessel color, contrast, and brightness, and a lesion-aware perturbation that selectively inpainted automatically detected lesions in non-vessel regions. The vessel, near-vessel, and optic-disc regions were preserved in both analyses.

Broad background perturbations substantially reduced classification performance relative to the FOV-standardized baseline accuracy of $0.806$. Accuracy decreased to $0.725$ after background color standardization, $0.763$ after contrast reduction, and $0.760$ after brightness standardization. Joint standardization of color, contrast, and brightness reduced accuracy to $0.682$, which was similar to the result obtained by replacing the valid non-vessel background with a shared constant value. These reductions indicate that RETFound was sensitive to broad variation in non-vessel appearance. However, because these regions may contain both clinically meaningful pathology and acquisition-related variation, the perturbations should not be interpreted as isolating acquisition bias alone.

To examine the contribution of localized lesion-related information more directly, non-vessel lesions were automatically detected using a pretrained fundus-lesion segmentation method~\cite{fundus_lesions_toolkit} and inpainted using LaMa~\cite{suvorov2022lama}. This lesion-aware perturbation produced only a small change in overall accuracy, from $0.806$ to $0.803$, and in DR recall, from $0.838$ to $0.831$. The limited change suggests that suppressing the automatically detected lesion regions alone did not substantially alter RETFound predictions in this evaluation. Nevertheless, it does not establish that RETFound was independent of lesion-related information, because lesions may have been incompletely detected and residual contextual or textural information may have remained after inpainting.

Overall, the larger effects of broad background perturbations than lesion-aware inpainting suggest that RETFound was sensitive to a mixture of non-vessel appearance characteristics rather than only to the localized lesions targeted by the inpainting procedure. These characteristics may include clinically relevant retinal abnormalities, local texture, illumination, reflections, pigmentation-related variation, and acquisition-dependent appearance. Detailed perturbation procedures and complete performance results are provided in Supplementary Section~S8 \textit{Controlled Non-Vessel Background Perturbations and Lesion-Aware Inpainting}.

\subsection{Qualitative Comparison with RETFound Explanations}
Although pretrained RETFound achieves strong classification performance, its learned image-level representations are less directly tied to predefined vascular measurements than the proposed ring-wise descriptors. The proposed method provides complementary interpretability by operating on explicit vascular features extracted from concentric annular regions. In this subsection, we qualitatively compare explanation maps derived from RETFound with those produced by our ring-based approach: RETFound provides pixel- or patch-level attribution maps, whereas the proposed method provides ring-level vascular evidence based on measurable retinal descriptors.

\subsubsection*{Ring-Based Evidence Heatmap}
\begin{figure}[!htbp]
\centering

\begin{minipage}[t]{0.42\linewidth}
\centering
\includegraphics[width=\linewidth]{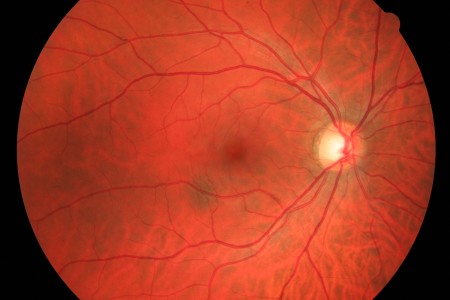}
\par\smallskip
\small (a) Original healthy fundus image.
\end{minipage}
\hfill
\begin{minipage}[t]{0.42\linewidth}
\centering
\includegraphics[width=\linewidth]{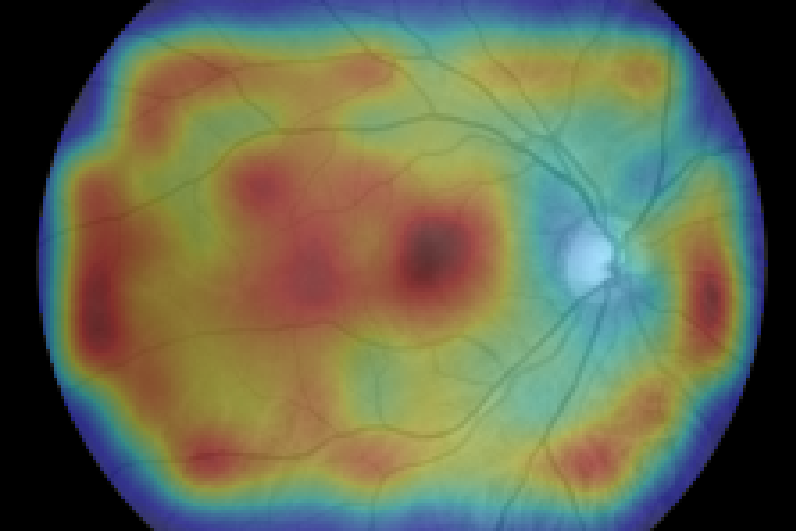}
\par\smallskip
\small (b) RETFound explanation map using patch-level Integrated Gradients.
\end{minipage}

\medskip

\begin{minipage}[t]{0.52\linewidth}
\centering
\includegraphics[width=\linewidth]{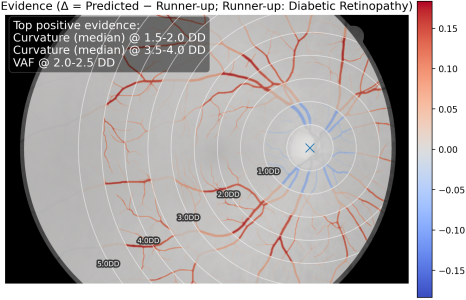}
\par\smallskip
\small (c) Proposed ring-based evidence heatmap.
\end{minipage}
\hfill
\begin{minipage}[t]{0.42\linewidth}
\centering
\includegraphics[width=\linewidth]{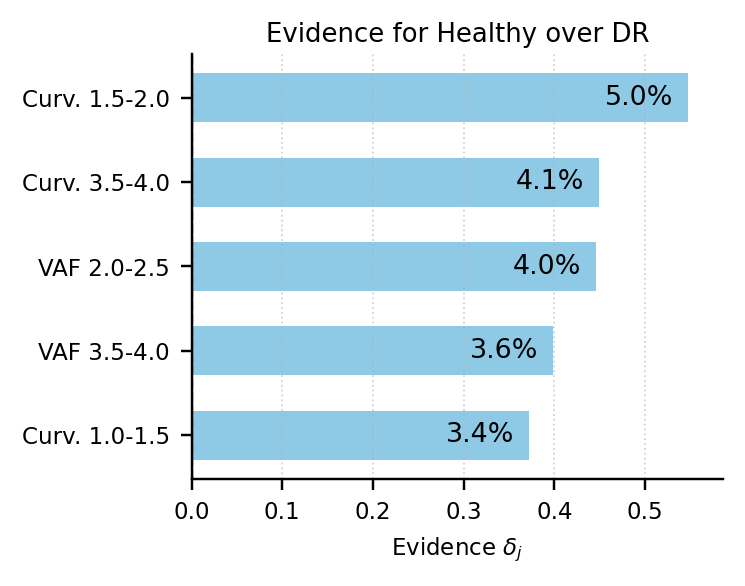}
\par\smallskip
\small (d) Top ring-wise feature evidence.
\end{minipage}

\caption{Qualitative and quantitative comparison of explanation maps for a representative healthy fundus image. Both methods correctly classify the image as healthy. The RETFound map shows patch-level positive Integrated Gradients attributions for the predicted-versus-runner-up logit margin. Warmer colors indicate stronger positive contributions to the model's preference for the predicted class over the runner-up class. The proposed heatmap shows signed evidence for the predicted class relative to the runner-up class ($\Delta=\mathrm{Predicted}-\mathrm{Runner\text{-}up}$), aggregated by ring and visualized on the retinal vasculature. Panel~(d) summarizes the top ring-wise feature contributions supporting the predicted class over the runner-up class.}
\label{fig:qualitative_comparison_01}
\end{figure}

Let $\mathbf{x}_{\mathrm{std}}$ denote the standardized feature vector (zero mean and unit variance per feature), and let $\mathbf{w}_c$ and $b_c$ denote the learned weights and bias for class $c$. The model produces class logits $\ell_c=\mathbf{w}_c^\top \mathbf{x}_{\mathrm{std}}+b_c$, yielding the predicted class $c^\ast$ and the runner-up class $\tilde{c}$ (the class with the second-largest logit). Each feature $j$ contributes a signed amount to the logit difference between $c^\ast$ and $\tilde{c}$. Specifically, we define the feature-level \emph{evidence} as $\delta_j=(w_{c^\ast,j}-w_{\tilde{c},j})x_{\mathrm{std},j}$, where $\delta_j>0$ indicates support for the predicted class over the runner-up and $\delta_j<0$ indicates support for the runner-up. Since the features are naturally grouped by annular rings, for a given ring mask $A$ we define the ring-wise evidence score as $\psi(A)=\sum_{j\in\mathcal{J}(A)}\delta_j$, where $\mathcal{J}(A)$ denotes the set of features extracted from ring $A$. The final heatmap assigns the ring-level score $\psi(A)$ to vessel pixels within ring $A$. Fig.~\ref{fig:qualitative_comparison_01} shows representative examples of this visualization. Rings with larger positive $\psi(A)$ appear more prominently, indicating stronger support for the predicted class over the runner-up.

\subsubsection*{RETFound Explanation Map}

Because transformer attention weights characterize internal token interactions rather than directly providing class-specific attribution to the final classifier output, we use Integrated Gradients (IG) \cite{sundararajan2017axiomatic} to explain the classification decision through the complete encoder--classifier pipeline. IG is computed for the logit margin between the predicted and runner-up classes by integrating its gradients over inputs linearly interpolated between a reference image corresponding to the ImageNet channel-wise mean color and the actual input image. The resulting signed pixel-wise attributions are converted to a non-negative support map by retaining positive values and summing across the RGB channels. To reduce noise, we apply SmoothGrad-IG \cite{smilkov2017smoothgrad} by averaging raw signed attributions from 12 noisy realizations of the input. Finally, the support values are averaged within each $16\times16$ patch to form a $14\times14$ patch-level map (\texttt{igpatch}), which is upsampled to $224\times224$, masked to the valid retinal field of view, and normalized within this region.

\subsubsection*{Comparison}
Fig.~\ref{fig:qualitative_comparison_01} compares the RETFound patch-level IG map (\texttt{igpatch}) with the proposed ring-based explanation for a representative healthy example. In this example the RETFound attribution map assigns positive attribution to broad image regions that are not closely aligned with the retinal vasculature, including the darker macular region around the fovea. This qualitative pattern is consistent with the controlled background-appearance experiments in Supplementary Section~S8, where perturbing non-vessel background appearance reduced RETFound performance. These results suggest that RETFound is sensitive to non-vascular image cues under the tested setting, although the attribution map itself should be interpreted as qualitative evidence. In contrast, the proposed explanation is explicitly grounded in annular vascular regions and ring-wise vascular descriptors. The heatmap identifies which annular regions provide positive evidence for the predicted class relative to the runner-up class, while the feature-evidence panel summarizes the top contributing descriptors. This provides a quantitative link between the model decision and measurable retinal vascular phenotypes, such as vessel curvature, regional vessel area fraction, caliber-related descriptors, branching density, and vascular connectivity. The feature-level attribution context analysis in Supplementary Section~S11, \textit{Feature-Level Reference Context for Ring-Wise Evidence}, further compares the raw values of the top evidence features with predicted-class and runner-up-class reference distributions. This provides dataset-level context for how these model-based vascular descriptors differ between the predicted and competing classes. These features should therefore be interpreted as model-based vascular evidence rather than standalone clinical diagnostic signs. Systematic clinician evaluation of explanation quality remains an important direction for future work.

\subsubsection*{Image-Level Clinical Interpretation}
\begin{figure}[!htbp]
\centering
\includegraphics[width=\linewidth]{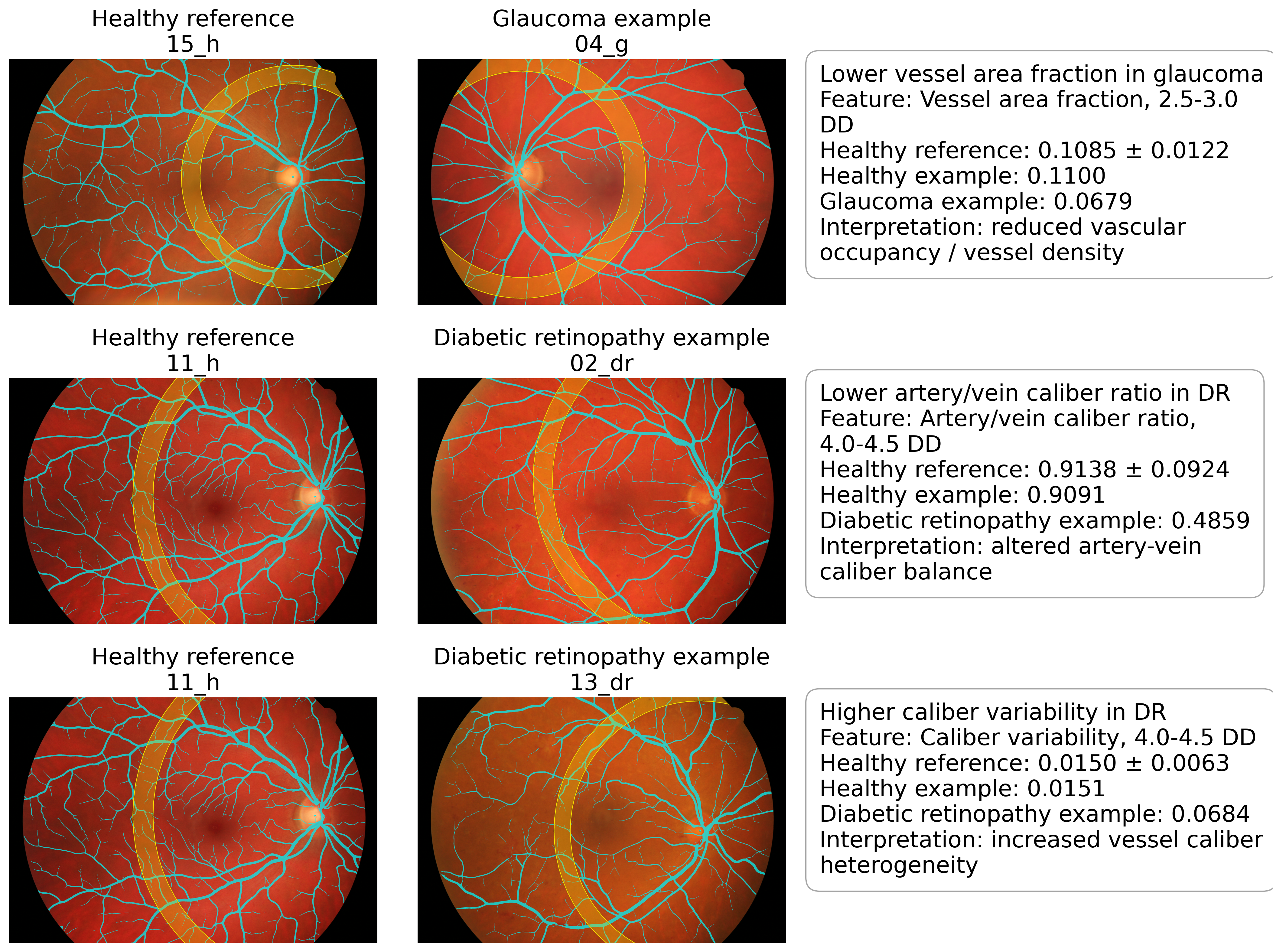}
\caption{
Representative image-level examples linking selected ring-wise vascular descriptors to clinically relevant retinal vasculature abnormalities. Cyan indicates the vessel mask, and the yellow band indicates the highlighted optic-disc-centered annulus. Each row compares a disease example with a healthy reference image for the same feature and annular region. The examples show reduced vessel area fraction in glaucoma, reduced artery--vein caliber ratio in diabetic retinopathy, and increased caliber variability in diabetic retinopathy. Feature values are compared with the healthy reference distribution for the same ring and descriptor.
}
\label{fig:ring_feature_clinical_examples}
\vspace{-0.1in}
\end{figure}
To further connect the extracted ring-wise descriptors with clinically relevant retinal vascular changes, we provide representative image-level examples in Fig.~\ref{fig:ring_feature_clinical_examples}. Each example shows a healthy reference image and a disease image with the vessel mask overlaid and the corresponding optic-disc-centered annulus highlighted. The selected feature value from the disease image is compared with the healthy reference distribution for the same descriptor and annular region. The examples include reduced vessel area fraction in glaucoma, reduced artery--vein caliber ratio in diabetic retinopathy, and increased caliber variability in diabetic retinopathy. These visualizations show how localized vascular deviations from healthy reference patterns can be inspected directly within the proposed ring-wise framework. The example-level interpretations are broadly consistent with prior retinal vascular studies summarized in Supplementary Section~S16, \textit{Related Work}, including reports of reduced vessel density in glaucoma and altered vessel caliber or caliber heterogeneity in diabetic retinopathy.

\subsection{Comparison Across Different Segmentation Algorithms}
\label{sec:segmentation_algorithms}

To assess the sensitivity of the proposed ring-based classification framework to the choice of vessel segmentation algorithm, we compared three segmentation methods: RIP-AV \cite{Dai2024}, FR-UNet \cite{Liu2022}, and SA-UNetv2 \cite{guo2025saunetv2}. The detailed segmentation implementation is described in Supplementary Section~S13 \textit{Feature-Family Importance Analysis}. Table~\ref{tab:segmentation_method_comparison} summarizes the classification results obtained using vessel masks generated by the three segmentation algorithms. RIP-AV achieved the highest accuracy and macro F1 on HRF, while on FIVES it performed similarly to SA-UNetv2 and better than FR-UNet. All three algorithms yielded similar results on SUSTech-SYSU.
\begin{table*}
\centering
\footnotesize
\caption{Comparison of the proposed ring-based method using vessel masks obtained from three segmentation algorithms: FR-UNet \cite{Liu2022}, SA-UNetv2 \cite{guo2025saunetv2}, and RIP-AV \cite{Dai2024}. Results are reported as mean accuracy, macro F1, and 95\% confidence intervals.}
\label{tab:segmentation_method_comparison}
\setlength{\tabcolsep}{2.5pt}
\begin{tabular}{lccccccccc}
\toprule
\multirow{2}{*}{\shortstack{\textbf{Segment.}\\\textbf{Method}}}
& \multicolumn{3}{c}{\textbf{HRF}} 
& \multicolumn{3}{c}{\textbf{FIVES}} 
& \multicolumn{3}{c}{\textbf{SUSTech-SYSU}} \\
\cmidrule(lr){2-4} \cmidrule(lr){5-7} \cmidrule(lr){8-10}
& Acc. & F1 & 95\% CI 
& Acc. & F1 & 95\% CI 
& Acc. & F1 & 95\% CI \\
\midrule
FR-UNet      
& 0.844 & 0.844 & [0.712, 0.923] & 0.685 & 0.653 & [0.637, 0.730] & \textbf{0.896} & \textbf{0.888} & \textbf{[0.876, 0.914]} \\

SA-UNetv2      
& 0.800 & 0.800 & [0.662, 0.891] & 0.722 & 0.695 & [0.675, 0.764] & 0.893 & 0.884 & [0.872, 0.910] \\

RIP-AV       
& \textbf{0.911} & \textbf{0.909} & \textbf{[0.793, 0.965]} & \textbf{0.724} & \textbf{0.700} & \textbf{[0.677, 0.766]} & \textbf{0.896} & \textbf{0.888} & [0.875, 0.913] \\
\bottomrule
\end{tabular}
\vspace{-0.2cm}
\end{table*}

Supplementary Figures~S23--S26 show the feature-importance rankings for feature groups aggregated across all annular rings under the three automatic segmentation settings and, where available, the corresponding ground-truth vessel mask settings. Across the segmentation-based settings, appearance-related feature groups, including vessel color, SO$_2$ proxies, and vessel--background entropy, were generally more important than geometric feature groups. In contrast, on the high-quality HRF dataset with ground-truth vessel masks, the top-ranked features included a mixture of geometric and appearance-related features. On the lower-quality FIVES dataset, however, the highest-ranked features remained predominantly appearance-related even when ground-truth masks were used. This pattern suggests that, when image quality is lower or vessel delineation is less reliable, appearance-related cues may remain more stable than fine-grained geometric measurements, even with expert vessel annotations. Overall, these findings suggest that the performance of the proposed framework depends in part on vessel-mask quality and segmentation characteristics, underscoring the importance of accurate vascular delineation for downstream ring-based classification. They also suggest that improved segmentation of fine vasculature may further enhance downstream classification performance.

\subsection{Ablation Study}

Using vessel masks predicted by RIP-AV on HRF, the complete feature representation achieved an accuracy of $0.911$. Removing the
SO$_2$-proxy features produced the largest performance decrease, reducing accuracy to $0.778$. Removing the vessel--background entropy features reduced accuracy to $0.800$, whereas removing either the geometry or red--green color features reduced accuracy to $0.844$.

This pattern differs from the standalone feature-family analysis in Supplementary Fig.~S27, in which the SO$_2$-proxy features achieved lower individual classification performance than the geometry and red--green color features. The two findings are nevertheless consistent because they measure different aspects of feature utility. Standalone evaluation measures the discriminative value of a feature family in isolation, whereas ablation analysis measures its incremental contribution when combined with
the remaining feature families. The results therefore indicate that the SO$_2$-proxy features provide complementary information that is not fully captured by the other feature families. Additional SO$_2$-proxy ablations across datasets are reported in Supplementary Section~S3.4.

\subsection{Cross-Dataset Validation}

Cross-dataset validation is challenging because retinal datasets may differ in imaging device, color response, field of view, resolution, preprocessing, and class composition. Such domain shifts can reduce external performance even when a model performs well within its source dataset. To evaluate external generalization, we trained the models on FIVES and evaluated them directly on HRF for three-class classification. No HRF images were used for model fitting, feature standardization, or model selection. For the proposed method, ground-truth vessel masks were used in both datasets to assess feature transfer independently of vessel-segmentation errors. Annular regions were constructed using a width of $0.5\times\mathrm{DD}$ and eight rings.

To reduce differences in spatial scale, each image was resized so that the optic disc had the same diameter in pixels while preserving the original FOV layout. Because RGB-derived descriptors may be sensitive to color-domain shift, we evaluated two configurations of the proposed representation. The first excluded both red--green color and SO$_2$-proxy features. The second retained the same structural descriptors and additionally included a within-image normalized SO$_2$-proxy descriptor. For each image, the median SO$_2$-proxy value was first computed within each ring. The resulting ring-wise values were then standardized by subtracting their within-image mean and dividing by their within-image standard deviation. This normalization reduces image-level offsets while preserving relative variation across retinal eccentricity.

Table~\ref{tab:cross_dataset_fives_hrf} summarizes the FIVES-to-HRF external validation results. The pretrained image-based baselines showed limited transfer, with accuracies ranging from $0.333$ to $0.356$. This finding should be interpreted within the present experimental setting and does not imply that pretrained retinal or natural-image encoders are intrinsically unsuitable for cross-domain retinal analysis. Rather, it indicates that frozen whole-image embeddings combined with a shallow classifier did not transfer effectively between these two datasets, potentially because of differences in image acquisition, color response, FOV geometry, resolution, preprocessing, or dataset composition.

The proposed ring-based representation showed stronger external performance. After excluding red--green color and SO$_2$-proxy features, it achieved an accuracy of $0.733$, a macro-F1 of $0.731$, and a macro ROC-AUC of $0.844$. Adding the within-image normalized SO$_2$-proxy descriptor increased accuracy to $0.778$ and macro-F1 to $0.780$, with a macro ROC-AUC of $0.849$. These results suggest that disc-normalized ring-wise vascular descriptors captured structural information that transferred more effectively in this specific FIVES-to-HRF setting, while relative ring-wise SO$_2$-proxy patterns provided additional complementary information.

The results also support a cautious interpretation of RGB-derived appearance descriptors. As shown in Supplementary Section~S3.2, \textit{Dataset-Dependent Distribution of RGB Color and SO$_2$-Proxy Features}, raw red--green color and absolute SO$_2$-proxy descriptors exhibit clear dataset-level shifts between FIVES and HRF. Raw color descriptors were therefore excluded from the cross-dataset configurations, and the SO$_2$-proxy was included only after within-image normalization. Although this normalization reduces image-level offsets, the resulting descriptor remains derived from RGB intensities and may still be affected by camera response, white balance, pigmentation, and preprocessing. Future work should investigate calibrated color harmonization, domain adaptation, fine-tuning of pretrained encoders, and validation across larger multi-device datasets.

\begin{table}[!htbp]
\centering
\caption{FIVES-to-HRF external validation. Models were trained on FIVES and evaluated directly on HRF. Ground-truth vessel masks were used for the proposed method in both datasets to assess feature transfer independently of vessel-segmentation errors.}
\label{tab:cross_dataset_fives_hrf}
\setlength{\tabcolsep}{4pt}
\renewcommand{\arraystretch}{1.06}
\begin{tabular}{lcccc}
\toprule
\textbf{Model} & \textbf{Acc.} & \textbf{Macro-F1} & \textbf{ROC-AUC} & \textbf{95\% CI (Acc.)} \\
\midrule
ResNet-50 & 0.356 & 0.248 & 0.637 & [0.232, 0.502] \\
ViT-B/16 & 0.333 & 0.167 & 0.694 & [0.214, 0.479] \\
ConvNeXt-B & 0.356 & 0.211 & 0.596 & [0.232, 0.502] \\
RETFound & 0.333 & 0.167 & 0.524 & [0.214, 0.479] \\
\midrule
\shortstack[l]{Proposed without RGB color\\and SO$_2$-proxy features} & 0.733 & 0.731 & 0.844 & [0.590, 0.840] \\
\shortstack[l]{Proposed with normalized\\SO$_2$-proxy descriptor} & \textbf{0.778} & \textbf{0.780} & \textbf{0.849} & \textbf{[0.637, 0.875]} \\
\bottomrule
\end{tabular}
\end{table}

\section{Discussion}

This study introduces an interpretable ring-based vascular representation for fundus image classification. The framework organizes vessel geometry, vessel color, oxygenation-related appearance, and vessel--background entropy within anatomically defined annular regions centered on the optic disc. Across HRF, FIVES, and SUSTech-SYSU, these descriptors provided useful classification information while linking model predictions to measurable retinal vascular characteristics. The results also clarify how segmentation quality, image appearance, and dataset shift affect the representation.

Segmentation quality affected downstream classification differently across datasets. On HRF, accuracy ranged from 0.800 to 0.911 with automatic masks, whereas accuracy reached 1.000 with expert masks. By contrast, the three segmentation algorithms produced similar results on SUSTech-SYSU. Because the datasets differ in sample size, image characteristics, disease composition, and classification setting, the smaller performance variation observed on SUSTech-SYSU cannot be attributed to sample size alone. Nevertheless, the HRF results indicate that more accurate vascular delineation can improve downstream ring-based measurements and classification.

Ablation analysis further showed that the feature families provide complementary rather than interchangeable information. Removing the SO$_2$-proxy features produced the largest decrease in HRF accuracy, from 0.911 to 0.778, despite their weaker standalone performance than that of the geometry feature group. Standalone analysis evaluates a feature family in isolation, whereas ablation evaluates the information lost when that family is removed from the complete representation. The results therefore suggest that the SO$_2$-proxy descriptors capture variation not fully represented by geometry, vessel color, or vessel--background entropy. Removing vessel--background entropy also reduced performance, indicating that perivascular intensity and contrast differences contribute additional information. These findings support a multifamily representation.

Cross-dataset validation from FIVES to HRF provided preliminary evidence that anatomically normalized ring-wise descriptors may retain useful information under dataset shift. The frozen whole-image embedding baselines showed limited direct transfer, whereas the proposed representation achieved higher accuracy in this cross-dataset evaluation after raw color features were excluded. Adding a within-image-normalized SO$_2$-proxy descriptor improved accuracy from 0.733 to 0.778. This suggests that relative ring-wise attenuation patterns may be more transferable than absolute RGB-derived measurements when image-level color offsets are reduced. Optic-disc-based spatial normalization may also contribute by expressing vascular measurements relative to an anatomical reference.

However, these cross-dataset findings should be interpreted cautiously. The limited transferability of the frozen pretrained embeddings does not imply that pretrained image models are intrinsically unsuitable for cross-dataset retinal image analysis. Their performance may be improved through multi-device training or alternative classification heads. Similarly, the higher accuracy of the ring-based representation in this evaluation does not establish general superiority across institutions or imaging devices. Rather, anatomical normalization and selective exclusion or normalization of color-sensitive features may help reduce certain forms of dataset shift.

Field-of-view standardization and non-vessel-background perturbation experiments suggested that whole-image models may use information not directly related to retinal vascular pathology. Changes in RETFound performance after field-of-view and background standardization do not imply that the model is clinically unreliable. Rather, under the evaluated FIVES setting, predictions were partly associated with dataset-specific spatial or appearance characteristics, possibly because acquisition properties were correlated with class labels. Explicit vascular descriptors provide greater transparency because the measurements used by the classifier can be inspected directly, although robustness across acquisition settings requires further validation.

Several limitations should be acknowledged. First, the framework depends on reliable vessel segmentation, optic-disc localization, and stable color and intensity measurements. This dependence is particularly important for images affected by uneven illumination, blur, media opacity, compression artifacts, low vascular contrast, or severe pathology. Image-enhancement and contrast-normalization methods may improve vessel visibility, but they may also alter the color and optical-density relationships used by the appearance-based descriptors. Their effects should therefore be evaluated at both the segmentation and feature-extraction stages.

Second, the evaluation was based on publicly available datasets containing primarily healthy, diabetic retinopathy, and glaucoma images. These datasets do not provide harmonized information about camera systems, acquisition protocols, screening environments, ethnicity, fundus pigmentation, or relevant clinical covariates. Moreover, the FIVES analysis used images selected according to the available image-quality labels. The present results therefore do not fully establish robustness across institutions, populations, imaging devices, or routine screening conditions. Validation using larger multi-institutional datasets with broader demographic and acquisition metadata, additional retinal diseases, and prospective clinical cohorts will be necessary.

Third, the SO$_2$-proxy descriptors are not calibrated measurements of retinal oxygen saturation. They are relative appearance features derived from RGB fundus images, whose broadband channels are affected by camera spectral sensitivity, illumination spectra, pigmentation, compression, and proprietary image-processing pipelines. These descriptors should therefore be interpreted as oxygenation-related optical-density proxies rather than direct physiological measurements. Their complementary classification value is encouraging, but physiological interpretation will require validation using better-characterized spectral imaging systems or calibrated multispectral and dual-wavelength measurements.

Finally, the current implementation was designed for reproducible analysis rather than real-time deployment. Most of the computational cost arises from CPU-based ring-wise feature extraction, whereas logistic-regression training and prediction are lightweight. Future work will focus on accelerating and integrating the processing pipeline through vectorized and parallel feature computation, GPU acceleration, and automated quality control. Further directions include uncertainty-aware vessel segmentation, identification of low-confidence rings, device-aware color harmonization, broader external validation, and integration of explicit vascular descriptors with deep image embeddings.

Overall, this study demonstrates that anatomically organized ring-wise vascular descriptors can provide an interpretable and complementary representation for retinal fundus image classification. By making the underlying geometric and appearance-related measurements directly accessible for inspection, the framework supports more transparent retinal-image analysis than classification outputs alone. Although broader validation across institutions, imaging devices, populations, and clinical settings is required, the results support the potential of ring-structured vascular measurements for retinal phenotype characterization, disease classification, and future investigations of associations with clinical outcomes.

\section{Method} \label{sec:method}

\subsection{Datasets} \label{sec:dataset}

We evaluated the proposed method on three publicly available retinal fundus image datasets with differing image characteristics, disease categories, and availability of expert vessel annotations to assess its robustness across dataset settings.

The High-Resolution Fundus (HRF) dataset~\cite{budai2013robust,hrfWebsite} contains 45 color fundus images, including 15 healthy, 15 diabetic retinopathy (DR), and 15 glaucoma images. Each image has a resolution of $3504 \times 2336$ pixels and is accompanied by an expert-annotated vessel mask and a FOV mask.

The FIVES dataset~\cite{jin2022fives} contains 800 fundus images with a resolution of $2048 \times 2048$ pixels, manual pixel-level vessel annotations, and image-quality labels for illumination or color distortion, blur, and low contrast. Based on these labels, we selected high-quality images from three classes: healthy ($n=168$), glaucoma ($n=83$), and DR ($n=130$).

The SUSTech-SYSU dataset~\cite{lin2020sustechsysu} contains high-resolution color fundus images with a resolution of $2880 \times 2136$ pixels and was collected for DR analysis. We used 1,016 images, including 385 DR and 631 healthy images, to evaluate binary DR-versus-healthy classification.

\subsection{Framework Overview} \label{sec:framework}

The proposed framework consists of ring-based retinal vascular feature extraction followed by disease classification. Starting from binary vessel masks obtained from expert annotations or existing segmentation methods, we extract geometry, red--green color, and optical-density-derived SO$_2$-proxy features from the segmented vasculature. Vessel--background entropy features additionally incorporate information from nearby perivascular pixels.

To preserve spatial variation across retinal eccentricity, the FOV is partitioned into concentric annular rings centered on the optic disc. Because the major retinal vessels emerge from the optic nerve head and branch outward across the retina~\cite{kocabiyik2005morphometric}, the optic disc provides an anatomically interpretable reference for organizing vascular measurements. Disc-centered regions are also commonly used for vessel-caliber and peripapillary vascular assessment~\cite{sherry2002reliability,yarmohammadi2016octa}. Unless otherwise specified, each ring has a radial width of half an optic disc diameter ($0.5\times$DD). Using DD as the radial unit provides an image-relative anatomical scale that reduces dependence on absolute image resolution and scale. This setting balances spatial specificity with sufficient vessel support for stable feature estimation. Classification performance remained stable across alternative ring widths, as detailed in Supplementary Section~S1.

\begin{figure}[!htbp]
\centering
\includegraphics[width=\linewidth]{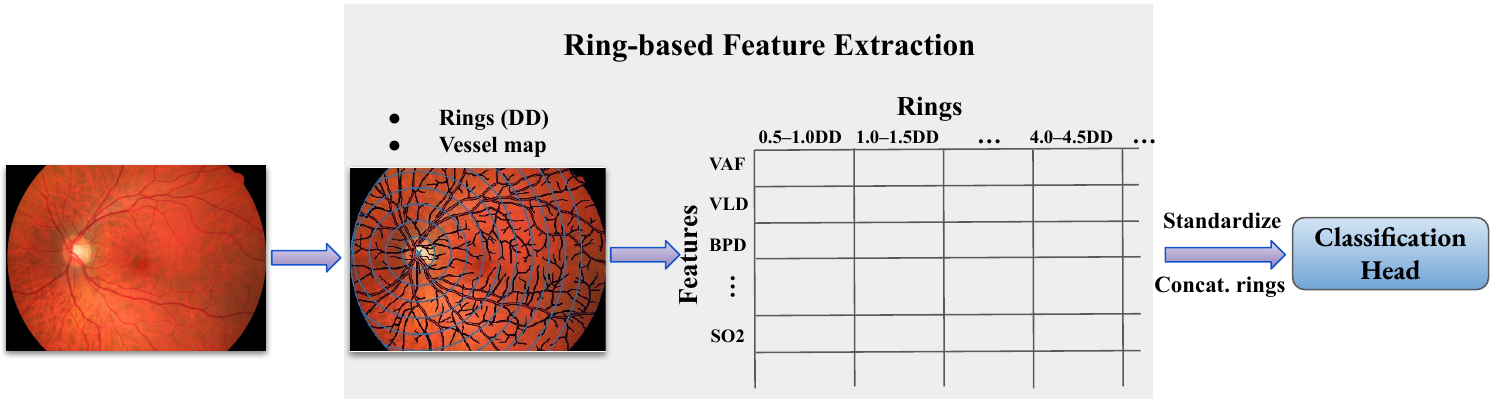}
\caption{Overview of the proposed ring-based classification framework. Concentric annular regions are defined around the optic disc, and vessel geometry, red--green color, OD-derived SO$_2$-proxy, and vessel--background entropy features are extracted within each ring. The resulting descriptors are concatenated, standardized, and classified using logistic regression.}
\label{fig:method}
\end{figure}

Within each ring, we compute four complementary feature families: vessel geometry, OD-derived SO$_2$-proxy appearance, red--green vessel color, and vessel--background entropy. The ring-wise descriptors are concatenated and standardized to account for differences in scale across feature types. The resulting feature vector is classified using logistic regression. Together, the anatomically organized representation and linear classifier permit direct examination of feature-level and annular-region contributions to individual predictions. Fig.~\ref{fig:method} summarizes the processing pipeline.

\subsection{Ring-Based Features}

The ring-based representation captures complementary aspects of vascular geometry, oxygenation-related appearance, red--green color, and vessel--background texture across retinal eccentricity. Pixel- and segment-level measurements within each ring are primarily summarized using the median and interquartile range (IQR), where $\mathrm{IQR}=Q_3-Q_1$ and $Q_1$ and $Q_3$ denote the 25th and 75th percentiles, respectively. The median represents the typical within-ring value, whereas the IQR quantifies spatial heterogeneity and is relatively robust to skewed distributions, segmentation noise, spurious vessel fragments, and focal atypical regions. Artery--vein contrast features are expressed as ratios of the corresponding median arterial and venous measurements. Inherently aggregated quantities, including vessel densities, counts, lengths, and graph statistics, are retained in their natural ring-level form. This spatial organization preserves regional variation between the peripapillary and more peripheral retinal vasculature. Supplementary Table~S4 summarizes the extracted features, their ring-wise statistics, and the rationale for each choice.

\subsubsection{Geometry-Based Vessel Features}

\paragraph{Vessel Density and Branching}

Vessel area fraction and vessel length density are established descriptors of retinal vascular  structure~\cite{Habib2014VascularGeometryDR,Dashtbozorg2016InfrastructureRIA,Andreini2025ArteriesVeins}. In this study,
we compute these descriptors within each annular region and additionally quantify branch-point density. Given an annular ring mask $A$, a binary vessel mask $V$, its one-pixel-wide skeleton $S$, and a branch-point mask $J$, vessel area fraction (VAF), vessel length density (VLD), and branch-point density (BPD) are defined as
\begin{equation}
\mathrm{VAF} = \frac{|A \cap V|}{|A|}, \qquad
\mathrm{VLD} = \frac{|A \cap S|}{|A|}, \qquad
\mathrm{BPD} = \frac{|A \cap J|}{|A|},
\label{eq:geometry_density_features}
\end{equation}
where $|\cdot|$ denotes the number of pixels and $\cap$ denotes set intersection. The skeleton $S$ is obtained from $V$ through morphological thinning to a one-pixel-wide centerline.

Because some datasets provide only a combined vessel mask without artery--vein labels, branch points are extracted directly from the complete skeletonized vascular network, and artery--vein crossings are not removed. Accordingly, BPD represents the junction density of the projected vascular network rather than a count of anatomically verified bifurcations. The branch-point mask $J$ is constructed by identifying skeleton pixels with more than two 8-connected neighbors (i.e., more than two neighbouring skeleton pixels among the eight pixels sharing an edge or corner with the current pixel), followed by removal of connected branch-point components containing fewer than three pixels to suppress spurious detections. A representative example is shown in Fig.~\ref{fig:vascular_feature_examples}(a).

\paragraph{Vessel Caliber and Related Thickness Features}

\textbf{Vessel caliber.}
Retinal vessel caliber is commonly estimated from segmented vessel boundaries or centerline-to-boundary distances
~\cite{niemeijer2011avr,Dashtbozorg2016InfrastructureRIA,ElhamiAsl2017Tracking}. In this study, for each skeleton pixel $p\in S_A$, where $S_A=A\cap S$ denotes the set of vessel-centerline pixels within annular region $A$, the local vessel diameter is estimated from the Euclidean distance transform (EDT) of the binary vessel mask:
\begin{equation}
d(p)=2\,\mathrm{EDT}(p),
\label{eq:local_diameter}
\end{equation}
where $\mathrm{EDT}(p)$ is the Euclidean distance, in pixels, from $p$ to the nearest vessel boundary. Multiplying this distance by two provides an approximation of the local vessel diameter under the assumption that the skeleton pixel lies near the vessel centerline. Within each ring, vessel-caliber variability is summarized by the IQR of the local diameter values $d(p)$. A representative visualization of this estimation is shown in Fig.~\ref{fig:vascular_feature_examples}(b).

\textbf{Artery--vein ratio.}
The artery--vein ratio is an established caliber-based descriptor of the relative widths of retinal arteries and veins
~\cite{sun2009retinal,niemeijer2011avr,Dashtbozorg2016InfrastructureRIA}. When artery and vein labels are available, the same local diameter estimates are used to calculate artery- and vein-specific median calibers. Let $S_A^{\mathrm{art}}$ and $S_A^{\mathrm{vein}}$ denote the arterial and venous skeleton pixels within ring $A$, respectively. The artery--vein ratio (AVR) is defined as
\begin{equation}
\mathrm{AVR}=
\frac{
\mathrm{median}\!\left(
d(p)\mid p\in S_A^{\mathrm{art}}
\right)}
{
\mathrm{median}\!\left(
d(p)\mid p\in S_A^{\mathrm{vein}}
\right)
}.
\label{eq:avr}
\end{equation}
This ratio characterizes the typical arterial caliber relative to the typical venous caliber within the same annular region. Median diameters are used to reduce sensitivity to local segmentation errors and atypical vessel segments.

\textbf{Thin-vessel length.}
When artery--vein labels are unavailable, we define two disc-normalized descriptors of relative vessel thickness: thin-vessel length and the thick-to-thin diameter ratio. Thin skeleton pixels are defined as
\begin{equation}
S_A^{\mathrm{thin}}
=
\left\{
p\in S_A:d(p)<\tau
\right\},
\qquad
\tau=\alpha DD,
\label{eq:thin_pixels}
\end{equation}
where $DD$ is the optic-disc diameter in pixels and $\alpha$ is a fixed proportional threshold. The thin-vessel length is then calculated as
\begin{equation}
L_{\mathrm{thin}}
=
\left|S_A^{\mathrm{thin}}\right|,
\label{eq:thin_length}
\end{equation}
where $|\cdot|$ denotes set cardinality. Because the skeleton is one pixel wide, $L_{\mathrm{thin}}$ represents the total thin-vessel centerline length within the ring in pixel units.

We set $\alpha=0.018$, corresponding to a diameter threshold of approximately $1.8\%$ of the optic-disc diameter. This value was selected empirically from the observed vessel-caliber distribution so that the threshold lies near the upper boundary of the small-caliber range. Scaling the threshold by $DD$ reduces its dependence on absolute image resolution and fundus scale. The resulting feature provides a ring-wise surrogate measure of small-caliber vessel content.

\textbf{Thick-to-thin diameter ratio.}
The complementary set of thick-vessel skeleton pixels is defined as
\begin{equation}
S_A^{\mathrm{thick}}
=
S_A\setminus S_A^{\mathrm{thin}}.
\end{equation}
The thick-to-thin diameter ratio is then calculated as
\begin{equation}
D_{\mathrm{ratio}}
=
\frac{
\mathrm{median}\!\left(
d(p)\mid p\in S_A^{\mathrm{thick}}
\right)}
{
\mathrm{median}\!\left(
d(p)\mid p\in S_A^{\mathrm{thin}}
\right)
}.
\label{eq:diameter_ratio}
\end{equation}
This feature quantifies the contrast between larger- and smaller-caliber vessels within the same ring. Together, $L_{\mathrm{thin}}$ and $D_{\mathrm{ratio}}$ provide artery--vein-label-free measures of small-vessel content and within-ring caliber heterogeneity. They should not be interpreted as direct substitutes for the anatomically defined AVR because the thick and thin groups do not necessarily correspond exactly to veins and arteries.

\paragraph{Vessel Tortuosity and Curvature}

\textbf{Vessel tortuosity.}
Vessel tortuosity is quantified using the arc-to-chord ratio, a commonly used centerline-based measure of the deviation of a vessel segment from a straight path~\cite{Owen2009CAIAR,ElhamiAsl2017Tracking,Huang2019}. For a segment represented by an ordered sequence of centerline points $\mathbf{p}_i=(x_i,y_i)$, $i=1,\ldots,N$, the cumulative arc length is defined as
\begin{equation}
\ell_1=0,\qquad
\ell_i=
\sum_{k=1}^{i-1}
\left\|
\mathbf{p}_{k+1}-\mathbf{p}_k
\right\|,
\quad i=2,\ldots,N,
\label{eq:cumulative_arc_length}
\end{equation}
where $\|\cdot\|$ denotes the Euclidean norm. The total arc length is $L=\ell_N$, and the chord length is
$C=\|\mathbf{p}_N-\mathbf{p}_1\|$. For segments with $C>0$, tortuosity is defined as
\begin{equation}
T=\frac{L}{C}\geq 1.
\label{eq:tortuosity}
\end{equation}
A value of $T=1$ corresponds to a straight segment, whereas larger values indicate greater deviation from the direct path between the segment endpoints. Within each annular region $A$, variability in segment tortuosity is summarized by the IQR.

\textbf{Curvature.}
Whereas the arc-to-chord ratio characterizes the overall shape of a vessel segment, curvature quantifies local bending along the centerline~\cite{ElhamiAsl2017Tracking,Huang2019}. In this study, the tangent orientation at an interior point $\mathbf{p}_i$ is approximated using a centered difference:
\begin{equation}
\theta_i=
\operatorname{atan2}
\!\left(
y_{i+1}-y_{i-1},
x_{i+1}-x_{i-1}
\right).
\label{eq:tangent_orientation}
\end{equation}
The magnitude of the local angular change per unit arc length is then estimated as
\begin{equation}
\kappa_{\theta}(\ell_i)
=
\left|
\frac{\Delta\theta_i}
{\ell_{i+1}-\ell_{i-1}}
\right|,
\qquad
\Delta\theta_i=
\operatorname{wrap}
\!\left(
\theta_{i+1}-\theta_{i-1}
\right),
\label{eq:discrete_curvature}
\end{equation}
where $\operatorname{wrap}(\cdot)$ maps angular differences to $(-\pi,\pi]$, preventing artificial discontinuities when the orientation crosses the $\pm\pi$ boundary. Because arc length is measured in pixels, $\kappa_{\theta}$ is expressed in inverse-pixel units. Within each ring, typical local vessel bending is summarized by the median curvature.

\paragraph{Vascular Connectivity Features}

Within each annular region $A$, the vessel skeleton is represented as an undirected graph $G_A=(N_A,E_A)$, where nodes correspond to skeleton pixels and edges connect pairs of 8-neighboring skeleton pixels. From this graph, we extract two connectivity descriptors: the vascular connection count, defined as the graph edge count $|E_A|$, and the mean vascular connectivity,
\begin{equation}
\mathrm{VC}_{\mathrm{mean}}
=
\frac{1}{|N_A|}
\sum_{n\in N_A}\deg(n)
=
\frac{2|E_A|}{|N_A|},
\label{eq:vascular_connectivity}
\end{equation}
where $N_A$ and $E_A$ denote the node and edge sets of the ring-restricted graph, respectively, and $\deg(n)$ is the number of graph edges incident to node $n$. The vascular connection count reflects the total number of pixel-level skeleton connections within the ring and therefore depends partly on the amount of vasculature present. Mean vascular connectivity normalizes this quantity by graph size and characterizes the average local connectedness of the skeleton.

These graph-based descriptors complement branch-point density (BPD). Whereas BPD measures the frequency of detected junctions relative to ring area, vascular connection count and mean vascular connectivity also describe the organization of neighboring skeleton pixels and may be sensitive to vessel continuity and fragmentation.

\subsubsection{OD-derived \texorpdfstring{SO$_2$}{SO2}-Proxy Features}

Prior studies have reported altered retinal oxygenation patterns in DR and glaucoma, including increased venous oxygen saturation in DR and reduced arteriovenous oxygen difference in glaucoma \cite{geirsdottir2012retinal,Olafsdottir2011GlaucomaOximetry}. Motivated by these findings, we include SO$_2$-proxy descriptors as one component of our ring-structured vascular representation.
\paragraph{Optical Density}
Our formulation is based on the dual-wavelength retinal oximetry principle, which combines a near-isosbestic green wavelength with an oxygen-sensitive red wavelength \cite{hammer2008retinal,beach1999oximetry,Hirsch2023}. Under a Beer--Lambert approximation, the optical density (OD) at wavelength $\lambda$ is
\begin{equation}
OD(\lambda) = -\ln\!\left(\frac{I_{\mathrm{in}}(\lambda)}{I_{\mathrm{out}}(\lambda)}\right)
\approx \varepsilon(\lambda)\,h\,l,
\label{eq:beer_lambert}
\end{equation}
where $I_{\mathrm{out}}(\lambda)$ and $I_{\mathrm{in}}(\lambda)$ denote the outside- and inside-vessel intensities, respectively, $\varepsilon(\lambda)$ is an effective haemoglobin extinction coefficient, $h$ is an effective haemoglobin concentration term, and $l$ is the effective optical path length. To reduce the influence of pigmentation and background reflectance, we use a local vessel-free reference estimated at the same spatial location as each vessel pixel.

In retinal vessels, haemoglobin is the dominant chromophore, and the OD spectrum can be approximated as the sum of oxy- and deoxyhaemoglobin contributions:
\begin{equation}
OD(\lambda) \approx
\varepsilon_{\mathrm{HbO_2}}(\lambda)\,h_{\mathrm{HbO_2}} +
\varepsilon_{\mathrm{Hb}}(\lambda)\,h_{\mathrm{Hb}},
\label{eq:od_two_species}
\end{equation}
where $h_{\mathrm{HbO_2}}$ and $h_{\mathrm{Hb}}$ are effective nonnegative coefficients that absorb the approximately wavelength-invariant path-length factor for a fixed vessel geometry. Green wavelengths near an isosbestic point are only weakly dependent on oxygen saturation, whereas red wavelengths provide stronger oxygen-sensitive contrast, motivating our RGB-based approximation.

\paragraph{RGB Optical Density Estimation and \texorpdfstring{SO$_2$}{SO2}-Proxy Inversion}
\label{sec:od_so2_estimation}
Because RGB cameras record band-integrated rather than monochromatic intensities, we model each channel $c\in\{R,G,B\}$ over nominal wavelength supports $\Lambda_B=[443,488]~\mathrm{nm}$, $\Lambda_G=[535,555]~\mathrm{nm}$, and $\Lambda_R=[600,620]~\mathrm{nm}$, following prior retinal oximetry practice \cite{Bartczak2016,Cameron2019DepthsCastShadow,Kollath2020DarkSkyUnit}, with each support discretized at 1-nm intervals. The observed intensity in channel $c$ at location $p$ is
\begin{equation}
\bar{I}_c(p) \;=\; \sum_{\lambda \in \Lambda_c} \Phi_c(\lambda)\,I(\lambda,p),
\label{eq:rgb_formation}
\end{equation}
where $\Phi_c(\lambda)$ denotes the effective spectral sensitivity of channel $c$ and $I(\lambda,p)$ is the reflected-light spectral signal reaching the sensor. Applying \eqref{eq:rgb_formation} to the vessel-free reference and vessel signal gives
\begin{equation}
\begin{aligned}
\bar{I}_{\mathrm{out},c}(p)
&\triangleq \sum_{\lambda\in\Lambda_c}
\Phi_c(\lambda) I_{\mathrm{out}}(\lambda,p),\\
\bar{I}_{\mathrm{in},c}(p)
&\triangleq \sum_{\lambda\in\Lambda_c}
\Phi_c(\lambda) I_{\mathrm{in}}(\lambda,p).
\end{aligned}
\label{eq:Iin_Iout_band_def}
\end{equation}

Under the Beer--Lambert approximation \eqref{eq:beer_lambert}, the channel-wise OD can be written as
\begin{equation}
OD_c(p)\approx -\ln\!\left(\frac{\bar{I}_{\mathrm{in},c}(p)}{\bar{I}_{\mathrm{out},c}(p)}\right)
\approx \bar{\varepsilon}_c\,h(p)\,l(p),
\label{eq:od_bar_eps}
\end{equation}
where $\bar{\varepsilon}_c$ is an effective channel-averaged extinction coefficient. Because $\bar{I}_{\mathrm{in},c}(p)$ and $\bar{I}_{\mathrm{out},c}(p)$ are measured under the same local illumination and camera settings, their ratio is less sensitive to multiplicative changes in brightness and smooth shading, although this may be degraded by nonlinear image processing, clipping, or wavelength-dependent system effects. This expression is a compact lumped approximation of haemoglobin attenuation over the band $\Lambda_c$. To make the channel-wise band averaging explicit, we introduce the corresponding local normalized within-band weighting,
$W_c(\lambda;p)=
\frac{\Phi_c(\lambda)\,I_{\mathrm{out}}(\lambda,p)}
{\sum_{\lambda' \in \Lambda_c} \Phi_c(\lambda')\,I_{\mathrm{out}}(\lambda',p)}$.
Assuming that, after local referencing, the normalized spectral shape of $I_{\mathrm{out}}(\lambda,p)$ varies only weakly across vessel locations within each channel band, we approximate $W_c(\lambda;p)$ by a channel-dependent weighting $W_c(\lambda)$. This yields
$\bar{\varepsilon}_c \triangleq \sum_{\lambda\in\Lambda_c} W_c(\lambda)\,\varepsilon(\lambda)$,
where $W_c(\lambda)$ represents the normalized effective within-band weighting for channel $c$.

The compact form in \eqref{eq:od_bar_eps} serves only as an intuitive lumped approximation. The final OD-derived SO$_2$-proxy inversion instead uses the species-resolved version of the same band-averaged model. Under the same within-band weighting,
\begin{equation}
OD_c(p)\approx
E_{c,\mathrm{HbO_2}}\,h_{\mathrm{HbO_2}}(p)
+
E_{c,\mathrm{Hb}}\,h_{\mathrm{Hb}}(p),
\label{eq:od_species_band}
\end{equation}
where, for haemoglobin species $s \in \{\mathrm{HbO_2}, \mathrm{Hb}\}$,
\begin{equation}
E_{c,s} = \sum_{\lambda \in \Lambda_c} \varepsilon_s(\lambda)\,W_c(\lambda).
\label{eq:Ecs_def}
\end{equation}
Here, $\bar{\varepsilon}_c$ denotes the lumped channel-averaged extinction coefficient in the simplified model, whereas $E_{c,s}$ denotes the species-resolved channel-averaged extinction coefficient used in the final inversion. Stacking the three RGB channels yields the matrix form
\begin{equation}
OD(p)\approx E\,\mathbf{h}(p),
\label{eq:od_linear_model}
\end{equation}
where
\[
\begin{aligned}
OD(p) &=
\big[\,OD_B(p),\,OD_G(p),\,OD_R(p)\,\big]^\top,\\
\mathbf{h}(p) &=
\big[\,h_{\mathrm{HbO_2}}(p),\,h_{\mathrm{Hb}}(p)\,\big]^\top .
\end{aligned}
\]
A detailed derivation of the RGB band-integrated OD model is provided in Supplementary Section~S2 \textit{Derivation of the RGB Optical-Density Approximation}.

\textbf{Effective within-band weighting estimation.}
Because camera spectral responses are typically unavailable for public RGB fundus datasets, the effective within-band weighting $W_c(\lambda)$ is approximated by a Gaussian-shaped band-pass profile:
\begin{equation}
\begin{aligned}
W_c(\lambda)
&= w_c \exp\!\left(
-\frac{(\lambda-\lambda_c)^2}{2\sigma_c^2}
\right),\\
w_c^{-1}
&= \sum_{\lambda'\in\Lambda_c}
\exp\!\left(
-\frac{(\lambda'-\lambda_c)^2}{2\sigma_c^2}
\right).
\end{aligned}
\label{eq:Wc_gauss}
\end{equation}
where $\lambda_c$ is the channel center wavelength and $\sigma_c$ controls the bandwidth, with $\mathrm{FWHM}_c = 2\sqrt{2\ln 2}\,\sigma_c$.

In our experiments, we set $\lambda_R=610$\,nm, $\lambda_G=545$\,nm, and $\lambda_B=465$\,nm. Unless otherwise stated, we use $\mathrm{FWHM}_B=45$\,nm, $\mathrm{FWHM}_G=20$\,nm, and $\mathrm{FWHM}_R=20$\,nm. Additional perturbation experiments on the red/green center wavelengths and channel bandwidths produced similar SO$_2$-proxy trends and class separation, suggesting robustness to moderate passband variation.

\textbf{Vessel-free background reference estimation.}
\label{para:vessel_free_reference}

To estimate the vessel-free reference $\bar{I}_{\mathrm{out},c}(p)$ at vessel locations, we construct a smooth background map separately for each RGB channel. The vessel mask is first dilated to define an exclusion region around the vasculature. This region is removed and filled using nearby non-vessel pixels with OpenCV's Navier--Stokes PDE-based inpainting method. Large-scale Gaussian smoothing is then applied to the inpainted image to obtain a smoothly varying vessel-free background estimate. The reference value
at each vessel pixel is obtained by sampling this background map at the same spatial location. Further implementation details and illustrative examples are provided in Supplementary Section~S4, \textit{Construction of the Vessel-Free Background Reference}, and
Supplementary Figs.~S9 and~S10.

\textbf{RGB extinction coefficients.}
Using Prahl's tabulated haemoglobin spectra~\cite{prahl_hemoglobin}, we compute the channel-averaged extinction coefficients for oxyhaemoglobin and deoxyhaemoglobin according to \eqref{eq:Ecs_def}. Stacking these coefficients yields the extinction matrix
\begin{equation}
E =
\begin{bmatrix}
E_{B,\mathrm{HbO_2}} & E_{B,\mathrm{Hb}} \\
E_{G,\mathrm{HbO_2}} & E_{G,\mathrm{Hb}} \\
E_{R,\mathrm{HbO_2}} & E_{R,\mathrm{Hb}}
\end{bmatrix}
\in \mathbb{R}^{3 \times 2}.
\end{equation}
In this RGB approximation, the green channel provides a near-isosbestic reference, the red channel provides oxygen-sensitive contrast, and the blue channel contributes additional short-wavelength information.

\textbf{OD-derived SO$_2$-proxy estimation.}
We estimate the effective haemoglobin coefficients using ridge-regularized least squares:
\begin{equation}
\hat{\mathbf{h}}(p)
=
\left(
E^\top E+\gamma_{\mathrm{reg}}\mathbf{I}_2
\right)^{-1}
E^\top OD(p),
\label{eq:ridge_inversion}
\end{equation}
where $\mathbf{I}_2$ is the $2\times2$ identity matrix and $\gamma_{\mathrm{reg}}>0$ is the ridge-regularization parameter.
A non-negativity constraint is enforced through element-wise clipping, $\hat{\mathbf{h}}(p)\leftarrow\max(\hat{\mathbf{h}}(p),0)$. The local OD-derived SO$_2$-proxy is then defined as
\begin{equation}
\mathrm{SO}_{2,\mathrm{proxy}}(p)
=
\frac{
\hat{h}_{\mathrm{HbO_2}}(p)
}{
\hat{h}_{\mathrm{HbO_2}}(p)
+
\hat{h}_{\mathrm{Hb}}(p)
}.
\label{eq:so2_like}
\end{equation}

\begin{figure}[!htbp]
\centering

\begin{minipage}[t]{0.45\linewidth}
\centering
\includegraphics[width=\linewidth]{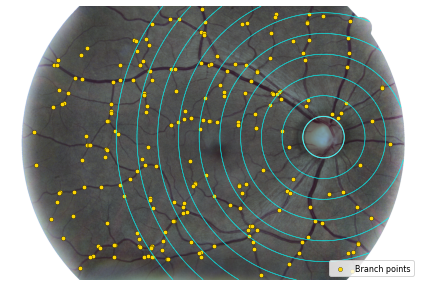}
\par\smallskip
\small (a) Branch-point extraction.
\end{minipage}
\hfill
\begin{minipage}[t]{0.45\linewidth}
\centering
\includegraphics[width=\linewidth]{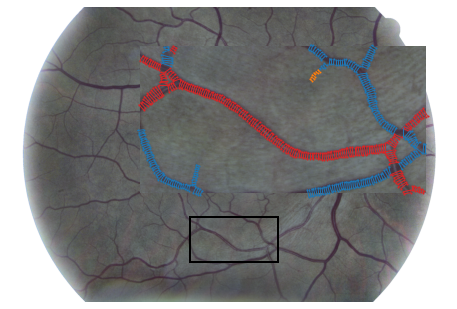}
\par\smallskip
\small (b) Vessel-diameter estimation.
\end{minipage}

\par\medskip

\begin{minipage}[t]{0.55\linewidth}
\centering
\includegraphics[width=\linewidth]{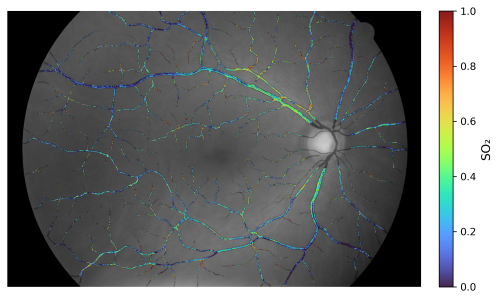}
\par\smallskip
\small (c) OD-derived SO\textsubscript{2}-proxy map.
\end{minipage}

\caption{Representative ring-wise vascular feature extraction. (a) Vessel branch points extracted from the skeletonized vasculature; yellow dots indicate skeleton pixels with more than two 8-connected neighbors. (b) Local vessel-diameter estimation along representative centerline segments. Red and blue indicate arterial and venous segments, respectively; orange indicates uncertain or unknown vessel labels; transverse markers indicate sampled local diameters. (c) Estimated OD-derived SO\textsubscript{2}-proxy values along the retinal vasculature of a representative healthy image.}
\label{fig:vascular_feature_examples}
\end{figure}

Because this RGB-based formulation relies on approximate passbands and does not explicitly model scattering or camera-specific image
processing, the resulting values are not calibrated physiological oxygen-saturation measurements. We therefore interpret 
$\mathrm{SO}_{2,\mathrm{proxy}}(p)$ as an oxygenation-related appearance descriptor. For each annular ring $A$, the proxy is evaluated on vessel pixels in $A\cap V$, and its within-ring variability is summarized by the IQR. When artery and vein masks are available, artery--vein contrast is represented by the ratio of the median arterial and venous SO$_2$-proxy values within the ring. A representative proxy map is shown in Fig.~\ref{fig:vascular_feature_examples}(c).

Because the proxy is estimated from RGB intensities rather than calibrated multispectral measurements, we additionally evaluated its
sensitivity to controlled global-brightness and channel-wise color-balance perturbations. The perturbation procedures and
corresponding results are provided in Supplementary Section~S3.3.




\subsubsection{Red--Green Color Features}

For vessel pixels $p\in A\cap V$, we compute two empirical red--green color descriptors from the original RGB image. The red-to-green ratio,
$I_R(p)/I_G(p)$, is summarized by its within-ring IQR to characterize variability in vascular color. The normalized red--green difference,
\[
\frac{I_R(p)-I_G(p)}{I_R(p)+I_G(p)},
\]
is summarized by its within-ring median to represent the typical chromatic contrast while reducing sensitivity to outliers. Here, $I_R(p)$ and $I_G(p)$ denote the red- and green-channel intensities at pixel $p$, respectively. Unlike the OD-derived SO$_2$-proxy features, these empirical descriptors are calculated directly from the original image intensities without background normalization or logarithmic transformation.

\subsubsection{Vessel--Background Entropy Features}

\paragraph{Fundus Image Preprocessing}

To obtain reliable vessel--background texture measurements, entropy features are computed from a photometrically normalized green-channel image, denoted by $I_G^{\mathrm{norm}}$. For each RGB channel, a smooth illumination field is estimated using large-scale Gaussian filtering. Ratio-based correction is then performed by dividing the original channel image pixel-wise by its corresponding illumination field, followed by linear rescaling. This procedure reduces low-frequency illumination variation and vignetting while preserving local retinal contrast.

The corrected RGB image is converted to CIE-Lab space, and contrast-limited adaptive histogram equalization (CLAHE) is applied to the luminance channel $L^*$. After conversion back to RGB, mild unsharp blending is applied to the corrected green channel to obtain $I_G^{\mathrm{norm}}$. The green channel is used because it generally provides strong vessel-to-background contrast in color fundus images. The complete preprocessing procedure and parameter settings are provided in Supplementary Section~S5, \textit{Photometric Normalization and Vessel-Enhancing Preprocessing}.

\paragraph{Image Inputs for Different Feature Groups}

This photometric-normalization pathway is used only for vessel--background entropy measurements. Red--green color features and OD-derived SO$_2$-proxy features are computed from the original RGB image to preserve the vessel-to-background intensity relationships required by these measurements. Geometry-based features depend only on the vessel masks and skeletons and are therefore independent of image normalization.

\paragraph{Local Entropy and Ring-Wise Statistics}

Local texture complexity is quantified using a Shannon entropy map computed from $I_G^{\mathrm{norm}}$. For each pixel $p$, entropy is calculated over a disk-shaped neighborhood $\mathcal{N}_r(p)$ with radius $r=5$ pixels:
\begin{equation}
H(p)
=
-\sum_{k:\,\pi_k(p)>0}
\pi_k(p)\log_2\!\bigl(\pi_k(p)\bigr),
\label{eq:local_entropy}
\end{equation}
where $\pi_k(p)$ denotes the empirical probability of gray level $k$ within $\mathcal{N}_r(p)$. In implementation, $I_G^{\mathrm{norm}}$ is clipped to $[0,1]$, rescaled to $[0,255]$, and converted to an 8-bit image with gray levels $k\in\{0,\ldots,255\}$. Because a base-2 logarithm is used, entropy is reported in bits.

Within each annular region $A$, entropy is summarized over two spatial supports: vessel centerline pixels and a near-vessel background band. Let $S$ denote the vessel skeleton. The near-vessel band $B_{\mathrm{near}}$ contains non-vessel pixels whose distance to the nearest vessel pixel lies between 2 and 6 pixels. The lower distance bound reduces direct vessel-boundary contamination, while the upper bound retains a localized perivascular background region.

The vessel-centerline and near-vessel background entropy features are defined as
\begin{equation}
H_{\mathrm S}(A)
=
\operatorname{median}
\left\{
H(p):p\in A\cap S
\right\},
\label{eq:vessel_entropy}
\end{equation}
\begin{equation}
H_{\mathrm{BG}}(A)
=
\operatorname{median}
\left\{
H(p):p\in A\cap B_{\mathrm{near}}
\right\}.
\label{eq:background_entropy}
\end{equation}
Here, $H_{\mathrm S}$ characterizes local texture complexity along vessel centerlines, whereas $H_{\mathrm{BG}}$ characterizes texture complexity in the immediately adjacent retinal background. These quantities are included as two separate classifier features; their difference is used only for descriptive interpretation. 

A descriptive analysis of the HRF dataset showed different vessel--background entropy patterns across classes. Healthy
images exhibited a moderate separation between vessel-centerline and near-vessel background entropy, with class-level median values of
$H_{\mathrm S}=3.849$ and $H_{\mathrm{BG}}=3.716$, corresponding to $\Delta H=0.133$. The separation was smaller for glaucoma
($H_{\mathrm S}=3.717$, $H_{\mathrm{BG}}=3.666$, and $\Delta H=0.051$) and was nearly absent for DR ($H_{\mathrm S}=3.871$, $H_{\mathrm{BG}}=3.870$, and $\Delta H=0.001$).

The reduced entropy separation in DR may reflect greater heterogeneity in the perivascular background associated with
abnormalities such as microaneurysms, hemorrhages, and exudates. The smaller separation in glaucoma may reflect subtler vascular or
perivascular appearance changes; however, this interpretation is less direct because glaucoma is primarily associated with alterations in the optic nerve head, retinal nerve fiber layer, and retinal vasculature rather than lesion-driven background abnormalities. These observations are descriptive, because local entropy is nonspecific and may also be affected by image quality and acquisition characteristics.

Representative ring-wise feature distributions across disease groups are shown in Supplementary Figs.~S16 and~S17, including geometry-based vessel features, vessel-appearance features, and vessel--background entropy features.

\subsection{Implementation and Evaluation Protocol}

For the proposed method, vessel masks were generated using RIP-AV~\cite{Dai2024}. For HRF and FIVES, which provide expert vessel
annotations, we additionally evaluated the proposed features using ground-truth vessel masks to quantify the influence of vessel-segmentation quality. Ring-wise vascular descriptors were concatenated to form one feature vector per image. Within each cross-validation fold, the standardization parameters were estimated from the training data and then applied to the held-out data.

The standardized feature vectors were classified using elastic-net-regularized logistic regression. The classifier configuration
was kept consistent across datasets, using the \texttt{saga} solver, an elastic-net penalty, an $l_1$ ratio of 0.10, a maximum of 3000 iterations, and class-balanced weights to reduce the influence of unequal class sizes. Only the regularization strength was dataset-specific, with $C=1.0$ for HRF and SUSTech-SYSU and $C=0.1$ for FIVES.

We compared the proposed ring-based representation with RETFound~\cite{zhou2023foundation}, a retinal foundation model pretrained
using a masked-autoencoder objective on approximately 1.6 million unlabeled retinal images. We used the publicly available color-fundus-photography checkpoint. To broaden the comparison, we additionally evaluated three ImageNet-pretrained image-classification backbones: ResNet-50~\cite{he2016resnet}, ViT-B/16~\cite{dosovitskiy2021vit}, and ConvNeXt-B~\cite{liu2022convnext}. For the frozen pretrained models, fundus images were resized to $224\times224$ pixels and normalized using the ImageNet channel-wise mean and standard deviation. The final image-level representation produced by each encoder was extracted and classified using the same elastic-net logistic-regression framework as the proposed features.

Lightweight fine-tuning was also evaluated within the training portion of each fold. For RETFound and ViT-B/16, the final two transformer blocks and the classification head were made trainable. For the convolutional backbones, the final two backbone stages and the classification head were made trainable.

Leave-one-out cross-validation was used for HRF and FIVES, whereas stratified five-fold cross-validation was used for SUSTech-SYSU. Accuracy, macro-F1, ROC-AUC, and per-class performance measures were calculated from the aggregated out-of-fold predictions.

Runtime profiling was performed using the high-resolution HRF images. Vessel segmentation and pretrained-encoder inference were conducted on a GPU, whereas ring-wise feature extraction, feature standardization, and logistic-regression prediction were conducted on a CPU. Detailed stage-wise runtimes and hardware specifications are provided in Supplementary Section~S17, \textit{Training and Inference Runtime Comparison}.

\subsection{Use of Generative Artificial Intelligence}

During the preparation of this manuscript, the authors used ChatGPT (OpenAI) to assist with improving the language, clarity, and organization of the manuscript. All AI-assisted text was critically reviewed, verified, and revised by the authors. The authors retained full responsibility for the scientific conception and development of the proposed framework, study design, conduct of the experiments, data analysis, interpretation of the results, conclusions, and final manuscript.

\section*{Data Availability}

The retinal fundus image datasets analysed in this study are publicly available from their original sources: the High-Resolution Fundus (HRF) dataset~\cite{budai2013robust,hrfWebsite}, the FIVES dataset~\cite{jin2022fives}, and the SUSTech-SYSU dataset~\cite{lin2020sustechsysu}. No new image datasets were generated during the current study.
\backmatter


\bibliography{sn-bibliography}

\section*{Acknowledgements}
This work was supported in part by an NSERC Discovery Grant (H.H.) and the Fields Institute. The authors thank Faisal Habib and other members of the Fields MDAS Lab for helpful and stimulating discussions.

\subsection*{Funding}
This study was funded by the Natural Sciences and Engineering Research Council of Canada (NSERC).

\section*{Author Contributions}

X.L. contributed to conceptualization, methodology, software development, investigation, validation, formal analysis, visualization, and writing the original draft. S.X. contributed to conceptualization, methodology, and reviewing and editing the manuscript. A.G. contributed to supervision, project administration, and reviewing and editing the manuscript. All authors read and approved the final manuscript.  H.H. contributed to conceptualization, methodology, supervision, project administration, and reviewing and editing the manuscript. All authors read and approved the final manuscript.

\section*{Competing Interests}
The authors declare no competing interests.

\end{document}